\documentclass{article}
\usepackage[table,dvipsnames]{xcolor} 
\usepackage{PRIMEarxiv}

\usepackage[utf8]{inputenc}
\usepackage[T1]{fontenc}

\usepackage{microtype}
\usepackage{graphicx}       
\graphicspath{{media/}}     
\usepackage{subfigure}
\usepackage{booktabs}       
\usepackage{fancyhdr}       
\usepackage{float}
\usepackage{tcolorbox}
\usepackage{algorithm}
\usepackage[noend]{algpseudocode}
\tcbuselibrary{listingsutf8}  
\definecolor{bordercolor}{RGB}{220,220,220} 
\definecolor{headingcolor}{RGB}{70,70,70}   
\definecolor{subheadingcolor}{RGB}{100,100,100} 
\definecolor{textcolor}{RGB}{0,0,0}         

\usepackage{hyperref}
\hypersetup{
    colorlinks=true,
    linkcolor=violet,
    citecolor=YellowOrange,
    urlcolor=Aquamarine
}

\usepackage{amsmath}
\usepackage{amsfonts}
\usepackage{algpseudocodex}
\usepackage{caption}       
\usepackage[framemethod=TikZ]{mdframed} 
\usepackage{mdframed}       
\usepackage{authblk}        
\usepackage[nameinlink]{cleveref}

\usepackage[numbers]{natbib}

\def\code#1{\texttt{#1}}                         
\def\methodname{Composite Learning Unit }         
\def\methodnameshort{CLU}                        
\usepackage[english]{babel}                      

\newcounter{problem} 
\renewcommand{\theproblem}{Problem~\arabic{problem}}

\newcommand{\problemref}[1]{\hyperref[#1]{\theproblem}}

\crefname{algorithm}{algorithm}{algorithms} 
\Crefname{algorithm}{Algorithm}{Algorithms}

\crefname{section}{section}{sections} 
\Crefname{section}{Section}{Sections}

\crefname{problem}{problem}{problems}  
\Crefname{problem}{Problem}{Problems}

\title{Composite Learning Units: Generalized Learning Beyond Parameter Updates to Transform LLMs into Adaptive Reasoners}

\author[1]{\href{https://www.linkedin.com/in/santoshkumarradha/}{Santosh Kumar Radha}}
\author[1]{\href{https://www.linkedin.com/in/oktay-goktas-8b2881167/}{Oktay Goktas}}
\affil[1]{Agnostiq Inc., 325 Front St W, Toronto, ON M5V 2Y1}
\affil[ ]{\texttt{contact@agnostiq.ai}}

\begin{document}

\maketitle
\vspace{-5pt}
\begin{abstract}
Human learning thrives on the ability to learn from mistakes, adapt through feedback, and refine understanding—processes often missing in static machine learning models. In this work, we introduce Composite Learning Units (CLUs) designed to transform reasoners, such as Large Language Models (LLMs), into learners capable of generalized, continuous learning without conventional parameter updates while enhancing their reasoning abilities through continual interaction and feedback. CLUs are built on an architecture that allows a reasoning model to maintain and evolve a dynamic knowledge repository: a General Knowledge Space for broad, reusable insights and a Prompt-Specific Knowledge Space for task-specific learning. Through goal-driven interactions, CLUs iteratively refine these knowledge spaces, enabling the system to adapt dynamically to complex tasks, extract nuanced insights, and build upon past experiences autonomously. We demonstrate CLUs’ effectiveness through a cryptographic reasoning task, where they continuously evolve their understanding through feedback to uncover hidden transformation rules. While conventional models struggle to grasp underlying logic, CLUs excel by engaging in an iterative, goal-oriented process. Specialized components—handling knowledge retrieval, prompt generation, and feedback analysis—work together within a reinforcing feedback loop. This approach allows CLUs to retain the memory of past failures and successes, adapt autonomously, and apply sophisticated reasoning effectively, continually \textit{learning from mistakes while also building on breakthroughs}.
\end{abstract}


Keywords


\section{Introduction}\label{sec:introduction}

\begin{figure}[!tb]
    \centering
    \includegraphics[width=\linewidth]{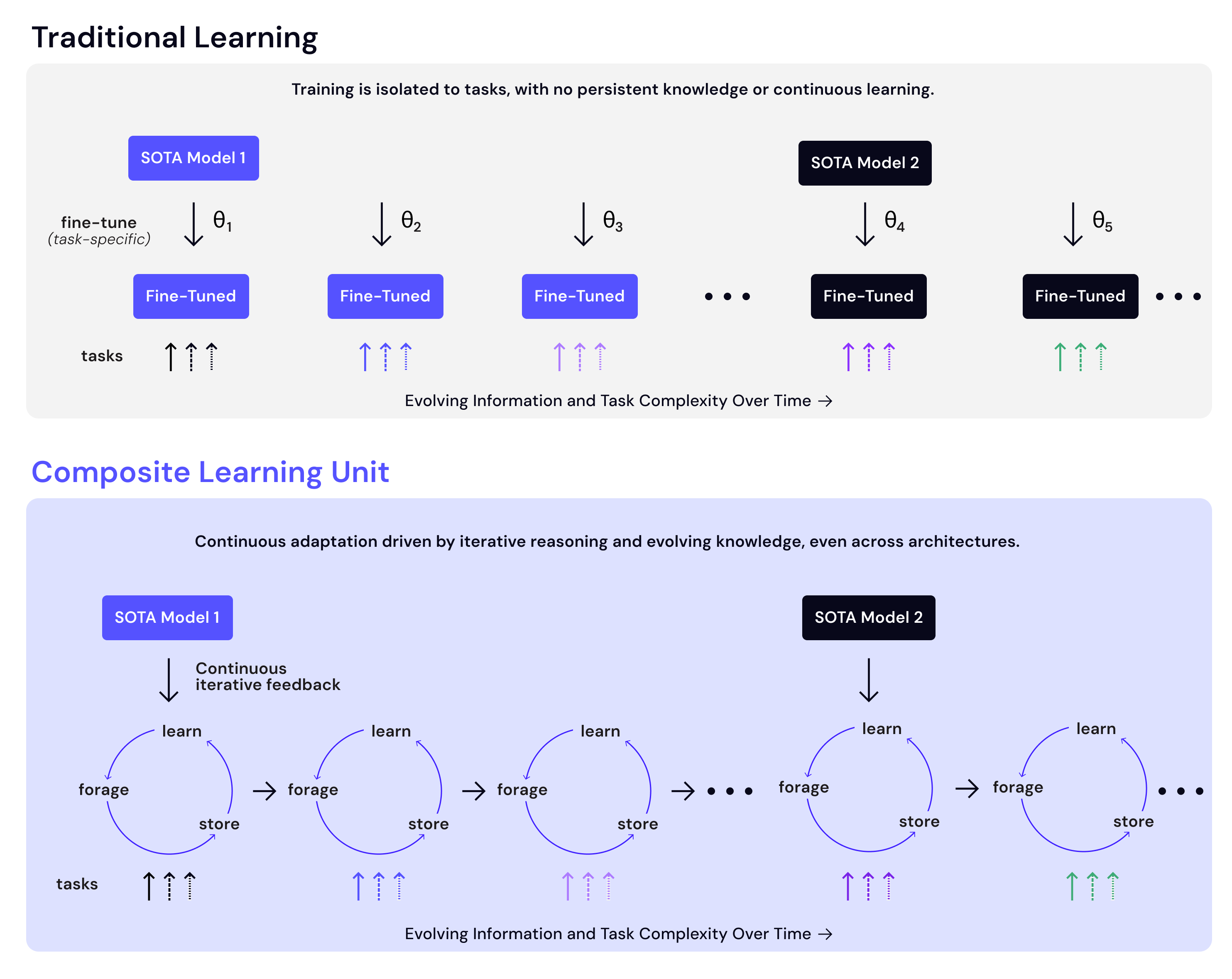}
    \caption{This illustration contrasts traditional learning methods, which rely on parameter updates and fine-tuning for each task, with Composite Learning Units (CLUs). By decoupling memory from reasoning, CLUs enable a feedback-driven continuous learning system that adapts iteratively, allowing the framework to evolve through reasoning and retain evolving knowledge for future tasks. This figure highlights how CLUs, driven by active inference, enable continuous adaptation and refinement beyond the limitations of static models.}
    \label{fig:hero}
\end{figure}

Artificial intelligence has seen transformative breakthroughs, most recently embodied by the emergence of Large Language Models (LLMs) such as GPT-4 \cite{openai2023gpt4}, Gemma \cite{team2024gemma}, LLaMA\cite{dubey2024llama}, Mistral \cite{jiang2023mistral7b}. These models, representative of transformer-based architectures, have demonstrated impressive capabilities in zero-shot reasoning \cite{brown2020language}, language generation, and problem-solving, often achieving near-human performance across a spectrum of specialized tasks\cite{radford2018improving,wei2022emergent,naveed2024comprehensiveoverviewlargelanguage,zhao2024surveylargelanguagemodels,vaswani2017attention,devlin2018bert}. However, they exemplify a fundamental limitation shared by most deep neural networks (DNNs): their reliance on static learning paradigms. Typically, such models require extensive retraining to adapt to new information, and even minor shifts in task requirements necessitate resource-intensive fine-tuning \cite{devlin2018bert}. This lack of adaptability is not limited to LLMs but pervades most modern DNNs, which are designed to extract knowledge from training data and encode it within static weight parameters. Online learning and continual learning methods have made strides towards overcoming these limitations by allowing incremental learning from new data and mitigating catastrophic forgetting \cite{gama2014survey,lopez2017gradient,robins1995catastrophic}; however, these approaches still often operate within predefined objectives, limiting their capacity for truly autonomous reasoning and adaptation to emergent complexities. Current learning processes in DNNs focus predominantly on mapping inputs to outputs—whether through classification, regression, or clustering—fitting models to recognizable patterns within data \cite{bishop2006pattern,lecun2015deep,schmidhuber2015deep}. This paradigm, while yielding substantial advances in supervised and unsupervised learning tasks, only addresses a limited aspect of the broader landscape of knowledge that can be inferred. Real-world scenarios demand more than pattern recognition; they require systems capable of extracting abstract, multifaceted, and evolving insights—capabilities that current static models, with their fixed objectives and representations, inherently lack \cite{sutton2019bitter,mccloskey1989catastrophic}. Consequently, models remain restricted in their ability to reason autonomously through complex latent properties or to discover emergent relationships beyond those explicitly presented during training\cite{fan2020addressing}. Thus, a critical question emerges: How can we design learning systems that transcend the limitations of static paradigms to actively discover, learn, and adapt to the myriad forms of knowledge embedded within data? 

To address this pressing challenge, recent research efforts have explored various methodologies aimed at enhancing neural networks' reasoning and adaptability. One prominent direction has focused on overcoming the limitations of fixed architectures by incorporating advanced reasoning frameworks into existing models. Chain of Thought and Multi-Thought\cite{wei2022chain,yao2024tree,radha2024iteration} techniques have extended the capabilities of transformer-based models by enabling stepwise dynamic problem-solving, thereby eliciting more sophisticated reasoning from pre-trained networks. These methods aim to guide models through complex tasks by breaking them down into smaller reasoning steps, allowing the models to leverage pre-learned associations in novel ways. Recent works have explored methods to distill these System 2 reasoning capabilities back into more efficient System 1 generations \cite{yu2024distilling}, aiming to improve performance without the computational overhead of intermediate reasoning steps. While these frameworks have succeeded in pushing the boundaries of what can be achieved with static models, they ultimately rely on augmentations rather than addressing the fundamental limitations of the static representations themselves. Current research efforts to overcome these limitations can be broadly classified into two categories: the development of fundamentally new architectures designed to provide inherent adaptability and incremental augmentation of existing models through increasingly complex system overlays \cite{lecun2022path,sahoo2024systematic}. The former aims to create new types of models inherently suited for dynamic learning, while the latter focuses on leveraging existing models, adding layers of complexity to achieve adaptability without reinventing the base model. Despite notable progress, these approaches still fail to facilitate truly autonomous, evolving learning. They remain tethered to predefined reasoning frameworks and static mechanisms for knowledge representation, which limits their ability to adapt dynamically and respond to changing contexts in real time. To overcome these fundamental constraints, we propose the Composite Learning Unit (CLU) — an evolving, feedback-driven architecture designed to transcend the static nature of traditional models. CLU utilizes a dynamic, agent-based approach that facilitates the continuous construction, refinement, and contextual adaptation of knowledge. By framing learning as an iterative, adaptive process, CLU bridges the gap between the fixed representations of conventional models and the need for real-time, evolving intelligence capable of autonomous, context-aware learning.

A key distinction of the CLU architecture is its ability to learn during inference. Traditional models, including transformer-based architectures, are limited by their static nature during inference, where they rely on pre-learned knowledge without adapting to new or changing inputs. In contrast, real-world tasks often demand dynamic adaptability, as task environments evolve or new information becomes available in real-time. This makes learning during inference not just a desirable feature but a critical necessity for systems operating in unpredictable, complex scenarios. The CLU system addresses this challenge by continuously refining its knowledge spaces through feedback gathered during task execution, allowing it to update its reasoning and improve performance without the need for costly retraining or parameter updates. By actively learning from each interaction, the CLU system ensures that it adapts and evolves as tasks unfold, offering a more flexible and responsive solution that transcends the static limitations of traditional models.

The \methodnameshort\ is designed on a foundation of three core principles drawn from philosophy, theoretical AI, and information theory: Constructivism\cite{sjoberg2010constructivism,chaput2004constructivist}, Active Inference\cite{parr2018anatomy,friston2016active,pezzulo2024active}, and Information Foraging\cite{pirolli2009elementary,pirolli1999information}. Constructivism underpins the concept that learning is an active construction process rather than a passive assimilation of information, enabling CLU to develop and refine complex relationships through ongoing task interactions and feedback. Active Inference extends this by emphasizing the system's proactive engagement with its environment, seeking information to reduce uncertainty and align with its goals, thereby ensuring that learning is not only reactive but inherently purposeful. Information Foraging adds a strategic dimension by optimizing the retrieval and utilization of knowledge, akin to an adaptive resource-gathering process, allowing the system to navigate its knowledge landscape efficiently. Together, these principles enable CLU to move beyond the limitations of fixed knowledge representations, providing a framework that supports the dynamic adaptation and construction of knowledge through continuous, goal-driven interaction with tasks and environments.

Moving beyond the conventional model-centric focus, CLU represents a paradigm shift towards an adaptive, multi-component learning architecture. Rather than being constrained by static weight parameters or extensive retraining requirements, CLU's design is centered on dynamic knowledge management. While our approach focuses on knowledge-based adaptation, other recent work has shown how Transformers can be trained to emulate and even improve upon traditional search algorithms like A* \cite{lehnert2024beyond}, demonstrating the potential for neural models to learn complex planning strategies. It achieves this through a multi-layered Knowledge Management System ($\mathcal{K}$), which comprises both a GKS ($\mathcal{K_G}$) for broad, reusable insights and a PKS ($\mathcal{K_P}$) that tailors knowledge to the unique requirements of individual tasks. These knowledge spaces are not static repositories but evolving entities that are continuously undergoing a process of refinement based on task outcomes and real-time feedback. By integrating a feedback loop into the learning process, CLU ensures that its internal knowledge and decision-making capabilities are always improving, adapting to changes in its operational environment without requiring resource-intensive retraining.

\begin{figure*}[ht]
    \centering
    \includegraphics[width=\linewidth]{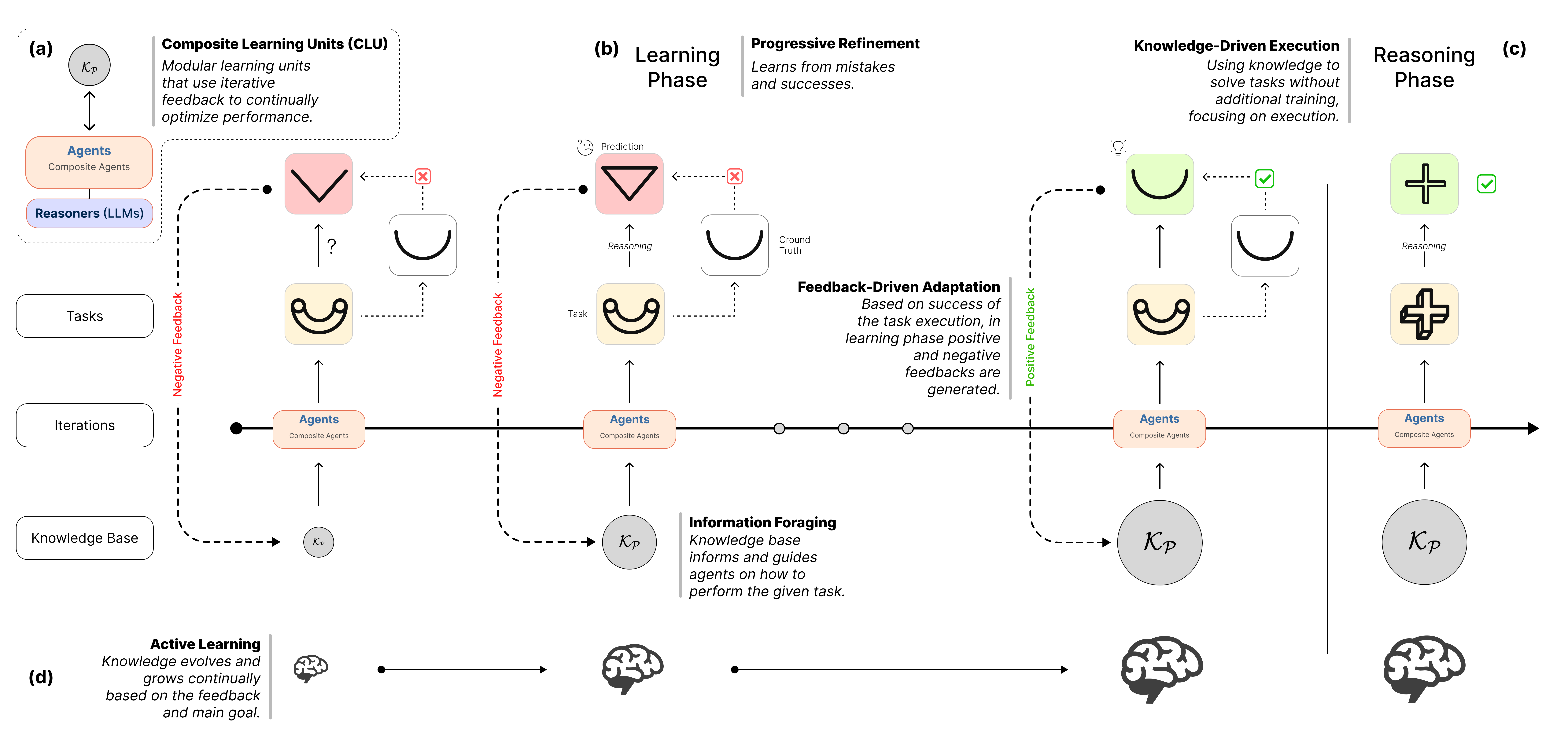}
    \caption{This figure shows the core components and dynamic processes of the Composite Learning Unit (CLU) framework, illustrating the adaptive process of CLUs, which are akin to intelligent systems that \textit{continually refine themselves through iterative practice and discovery to evolve and improve their reasoning abilities}. (a) CLUs are modular learning units that employ Large Language Models (LLMs) as base reasoners, leveraging feedback to optimize their performance iteratively. Agents, composed of these reasoners, adapt progressively through evolving knowledge bases that respond to tasks based on set goals. (b) The Learning Phase in the figure illustrates an example of learning a shape transformation task, where CLUs iteratively refine their understanding of the transformation rules. Given few examples, the unit learns through positive and negative feedback loops—progressively improving until it accurately understands and executes the transformation (c) The Reasoning Phase demonstrates how CLUs utilize accumulated knowledge to solve tasks effectively, focusing on execution without the need for additional training. (d) The knowledge base evolves through Active Learning, where knowledge grows via continuous practice and experience via feedback. See \cref{sec:theory-framework} for more details.}
    \label{fig:hero}
\end{figure*}

The overall architecture of CLU allows for fluid interaction between task-specific and general knowledge, facilitating prompt generation and task refinement through its dynamic operational mechanisms. This structure positions CLU to overcome the limitations faced by traditional neural networks, such as their reliance on pre-trained associations and fixed objectives. Instead of requiring augmentation to improve adaptability, CLU inherently supports a continuous learning and reasoning process. Each component within the system is designed to operate cohesively, refining knowledge and optimizing performance through a continuous feedback mechanism that aligns with changing goals and contexts.This iterative learning process, where CLUs continually refine themselves through iterative practice, is illustrated in \cref{fig:hero}. The figure showcases the core components and dynamic processes of the CLU framework, depicting how feedback loops guide the learning and reasoning phases to evolve and improve over time. \Cref{sec:theory-framework} will further elaborate on the foundational principles and interactions within CLU, highlighting how each part contributes to its adaptive learning capabilities.

Much like how modular systems leverage multiple components working in coordination—akin to multi-agent approaches\cite{guo2024large}—Composite Learning Units (CLUs) can be envisioned as adaptable building blocks within a broader, cooperative framework. The combined structure of multiple CLUs forms what we call a Composite Learning System (CLS), which is designed to tackle sophisticated problems by leveraging the collective strengths of individual learning units. A CLS serves as an overarching architecture in which multiple CLUs work together, each contributing its specialized capabilities to solve complex and evolving challenges. This cooperative setup enables the CLS to handle more demanding tasks by drawing on the adaptive learning abilities of its constituent CLUs. Although this paper focuses on the foundational concepts and adaptive mechanisms of a single CLU, the broader potential of the CLS approach, which involves integrating multiple CLUs for addressing more intricate tasks, is reserved for future exploration.

The remainder of this paper is structured as follows. In \cref{sec:theory-framework}, we provide a detailed theoretical foundation for the Composite Learning Unit (CLU), delving into the principles that underpin its adaptive capabilities and how its components interact cohesively. In  \cref{subsec:general-arcitecture} we take an in-depth look at the general architecture of CLU, outlining the operational dynamics and the role of its knowledge management units. In \cref{sec:KMU}, we describe the feedback mechanisms and the knowledge refinement processes that drive continuous learning within CLU. To illustrate the practical potential of this architecture, we provide a proof-of-concept demonstration in \cref{sec:results}, highlighting how CLU effectively adapts in a representative scenario in which it is exposed to a series of diverse tasks. Rather than an exhaustive benchmarking analysis, this example serves to validate the fundamental working principles and adaptive nature of our proposed system. Finally, \cref{sec:conclusion} concludes the paper with a summary of the contributions, key insights, and prospective avenues for future exploration, including extending CLUs into Composite Learning Systems (CLS) to tackle even more complex learning challenges.

\begin{figure*}[!htb]
    \centering
    \includegraphics[width=\linewidth]{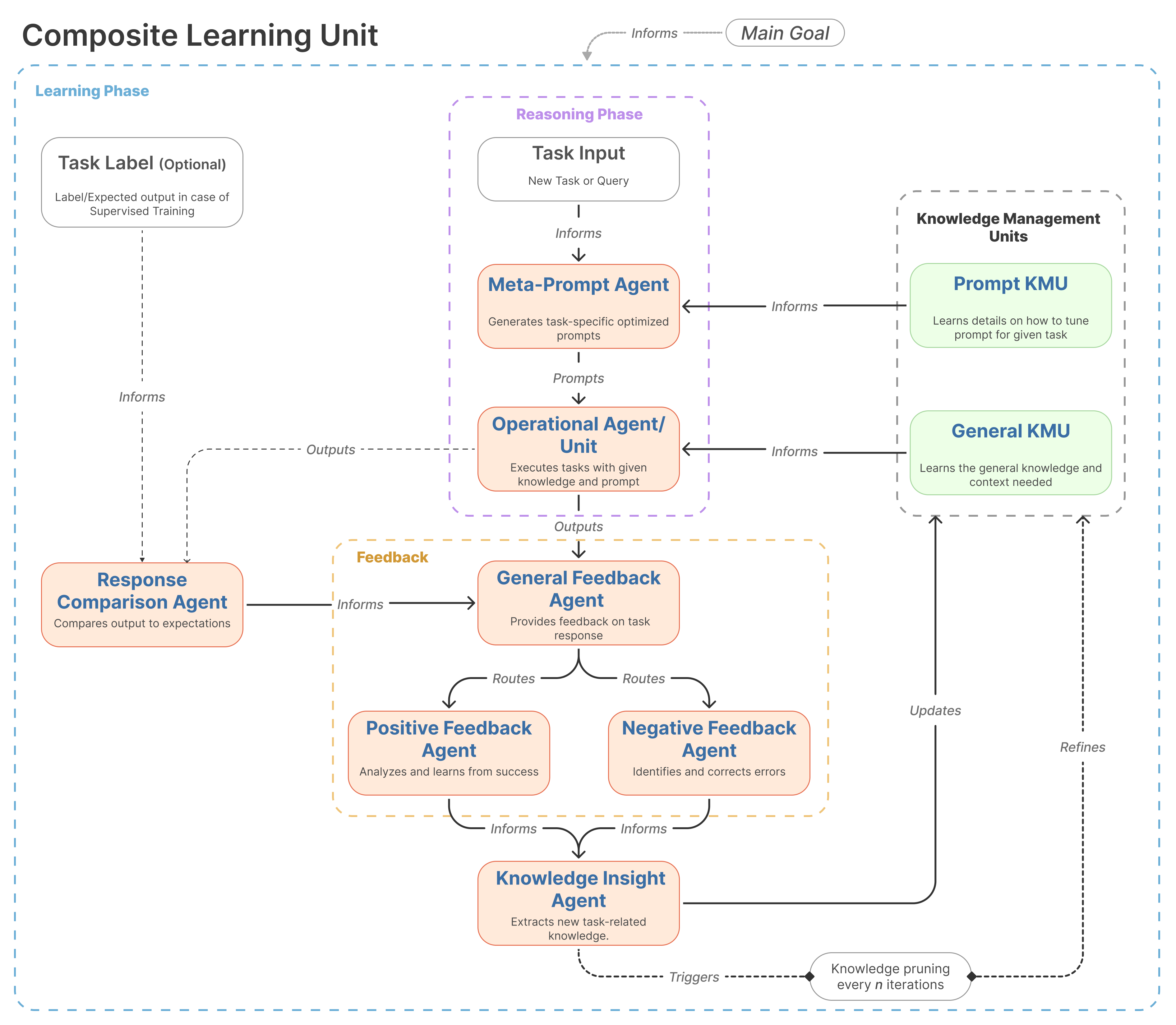}
    \caption{Process flow within the \methodname\ (\methodnameshort) framework. This diagram outlines the interaction between different agents, task inputs, and knowledge spaces, illustrating how feedback is integrated to improve the system's performance over time. 
The CLU framework operates in two distinct phases: the \textit{learning phase}, where the system iteratively refines its internal knowledge representations based on feedback from diverse datasets, and the \textit{reasoning phase}, where the system applies its existing knowledge to solve tasks without altering its internal state. Details about the components and their interactions are discussed in \cref{sec:theory-framework}.}
    \label{fig:clu}
\end{figure*}

\section{Theoretical Foundations and System Architecture}\label{sec:theory-framework}

The goal of the Composite Learning Unit (CLU) framework is to develop a black-box system composed of interacting agents that function as a general learning unit. This unit is designed to handle a wide range of tasks and datasets, whether supervised or unsupervised, and adaptively learn from the given data. By ``learning,'' we refer to the ability of the CLU to extract meaningful patterns, relationships, and insights from the data in pursuit of specific abstract goals, which may vary across different tasks. For instance, these goals may range from performing standard classification tasks to understanding higher-level abstract relationships between data points, such as interactions among different labels in a dataset. The CLU is structured to accommodate these varying objectives, allowing for dynamic task execution through a cooperative set of agents that continually adapt to new information.

The CLU framework generalizes the learning process by introducing a system that can adapt both to the nature of the data and the specific goal assigned to the task. The core problem it addresses is how to enable a learning unit to effectively process arbitrary goals and diverse datasets while maintaining the flexibility to learn from ongoing feedback and update its knowledge representations accordingly. This adaptability is captured in two key phases of the CLU's operation: the \textit{learning phase} and the \textit{reasoning phase}. Unlike traditional machine learning models, where the term ``training'' refers to the adjustment of weights through backpropagation, the \textit{learning phase} of CLU refers to the iterative refinement of its internal knowledge spaces based on feedback. In contrast, the \textit{reasoning phase} is akin to what is commonly called inference, where CLU utilizes its current knowledge to solve given tasks without modifying its internal state. This distinction allows CLU to maintain generalization capabilities across diverse tasks without requiring conventional retraining. The expected outcome of the CLU framework is an operational system capable of dynamically generalizing across tasks, executing high-level reasoning, and evolving with changing datasets and objectives. 

In this section, we explore the general architecture and operational dynamics of the CLU framework. First, \cref{subsec:general-arcitecture} formally defines the problem, outlining the core components of CLU, including task objectives, goal specifications, and the evolving knowledge spaces that underpin the learning process. Next, \cref{subsec:task-agents} introduces the key agents responsible for task execution, detailing their interaction with general and prompt-specific knowledge to generate outputs. The subsequent sections, \cref{sec:KMU} and \cref{sec:feedback}, focus on the Knowledge Management Unit (KMU), which oversees dynamic knowledge alignment and retrieval, and the feedback mechanisms that continuously refine this knowledge. Finally, \cref{sec:OpDynamics} addresses the operational dynamics during both learning and reasoning phases, distinguishing their respective roles within the CLU framework.

\subsection{General architecture and definitions} \label{subsec:general-arcitecture}

We formally define the problem as follows: Let \( \mathcal{T} \) represent the task space, where each task \( t \in \mathcal{T} \) is characterized by an input space \( X_t \) and an output space \( Y \). The \methodnameshort\ aims to process the input data \( x \in X_t \) and generate an output \( y \in Y \) in alignment with a specific goal \( G \). The goal \( G \) is an abstract representation of what the system aims to learn from the task and could vary in complexity and scope.  It defines the nature of the task that the system is trying to solve. For example, $t$ could represent a classification task, a regression task, or an unsupervised clustering task for the given data. It encapsulates the abstract goal and a specification of what the system needs to learn from the data. This is the task-level instruction that informs the system on which aspects of the data to devote the most focus.

In addition to the task and goal, the CLU relies on two types of knowledge spaces, which act as dynamic, trainable memory stores that evolve over time. These knowledge spaces are crucial for adapting the learning process based on the task and feedback received. In neural networks, trainable weights evolve based on data to learn and extract meaningful features. Similarly, in the CLU, knowledge spaces dynamically develop to store essential latent representations, which are continuously refined through iterative feedback cycles to guide future tasks. As illustrated \cref{fig:clu}, the CLU’s knowledge spaces act as critical resources that interact with agents during both task execution and feedback processing. These knowledge spaces are capable of encompassing high-level insights, particularly tailored for language-based tasks and data, due to the use of language models as the reasoning components. However, they are inherently flexible enough to represent abstract latent information distilled from broader sensory inputs during reasoning or interaction with external systems. This versatility allows the CLU to adapt and respond effectively across a wide range of applications by leveraging its evolving knowledge repositories, while maintaining a text-based latent structure as a primary representation. The knowledge spaces can be formally defined as follows:

\begin{itemize}
    \item \textbf{GKS} \( \mathcal{K_G} \): General Knowledge Space (GKS) serves as a repository for domain-specific knowledge, capturing broader patterns and insights related to the overall goal \( G \). It stores knowledge that generalizes across tasks, forming the core reasoning foundation of the \methodnameshort. In more formal terms, the task $t$ defines the problem context for which the system will retrieve knowledge from the GKS. 
    \item \textbf{PKS} \( \mathcal{K_P} \): Prompt-Specific Knowledge Space (PKS) is designed to store task-specific information related to generating effective prompts for the current task. It is focused on learning detailed relationships within the task at hand, ensuring that the system can adapt to specific nuances or requirements of a given task.
\end{itemize}

The need for two distinct knowledge spaces arises from the need to disentangle the general reasoning process from task-specific adaptations. While the GKS \( \mathcal{K_G} \) provides a high-level understanding of how the task relates to the overall goal, the PKS \( \mathcal{K_P} \) fine-tunes the prompt generation process to adapt to task-specific data. Together, these two knowledge spaces work in tandem, allowing the \methodnameshort\ to solve complex tasks by integrating both task-specific insights and high-level goal-driven reasoning. The retrieved knowledge from these spaces is used to guide the operational decisions of the \methodnameshort\ as depicted in \cref{fig:clu}.

There are several agents within the \methodnameshort\ that operate on this knowledge base and data, making decisions to modify and update the knowledge spaces based on feedback received during task execution. The primary agent, known as the \textit{Operational Agent} \( A_O \) (more details in \cref{subsec:task-agents}) is responsible for processing the input data \( x \in X_t \), along with the general knowledge \( k_G \in \mathcal{K_G} \) and the prompt-specific knowledge \( k_P \in \mathcal{K_P} \), to generate the output \( y \in Y \). The goal-specific reasoning of the system is encapsulated by these agents, which dynamically adapt their operations based on feedback.

The formal learning objective of the \methodnameshort\ is to maximize its performance across the task space \( \mathcal{T} \) through an iterative refinement of both the general and PKSs. Specifically, the aim is to optimize the knowledge spaces, \( \mathcal{K_G} \) and \( \mathcal{K_P} \), such that the system effectively achieves the desired goal \( G \) from the input data \( X_t \). This optimization process can be framed as follows:

\begin{equation}
    \max_{\mathcal{K_G}, \mathcal{K_P}} \mathbb{E}_{t \sim \mathcal{T}} \left[ Q\left(A_O(x, R_G(\mathcal{K_G}, t), A_{MP}(t, R_P(\mathcal{K_P}, t)))\right) \right],
\end{equation}\label{eq:maximize_exp}

where \( Q(\cdot) \) represents a quality function that assesses the performance of the operational agent \( A_O \) in completing the task \( t \), given the specific goal \( G \). Here, \( A_{MP} \) refers to the \textit{Meta-Prompt Agent}, which generates prompts tailored to each task based on the knowledge retrieved from the PKS \( \mathcal{K_P} \). The retrieval functions \( R_G(\mathcal{K_G}, t) \) and \( R_P(\mathcal{K_P}, t) \) denote the extraction of the most relevant knowledge from the general and PKSs, respectively, ensuring both the overarching goal and task-specific details are adequately accounted for.

The formulation above does not explicitly define \( Q(\cdot) \), as this function encapsulates an evolving objective influenced by the feedback received during task execution. This subtle but crucial aspect of the CLU draws inspiration from reinforcement learning (RL) \cite{mnih2015human}, where the feedback loop serves as a reward mechanism that iteratively guides the learning trajectory. It should also be noted that our approach of iterative refinement shares conceptual similarities with recent work on iterative preference optimization for reasoning tasks \cite{pang2024iterative}, although our method focuses on knowledge refinement rather than direct optimization of reasoning steps. In CLU, the feedback derived from each task—whether positive or corrective—directly influences the refinement of the knowledge spaces, effectively acting as a dynamic reward signal that aims to improve the performance of the system across multiple tasks. This optimization objective reflects a continuous learning paradigm in which feedback provides the necessary signals to enhance the latent representations within the GKPs and PKSs iteratively. As a result, the CLU adjusts its knowledge repository in an adaptive manner, learning to improve on successive tasks by utilizing insights obtained from previous ones. Similar to value updates in RL, the feedback-based mechanism employed here iteratively aligns the knowledge spaces to optimize the expected performance of the operational agent, ensuring an evolving and adaptive response to complex and dynamic tasks. Unlike rewards in RL, which provide only a final reward signal and are challenging to design\cite{sowerby2022designing} precisely due to their abstract nature, the feedback in CLU conveys reasoning about successes and failures, thereby adding explicit knowledge rather than abstract signals, enriching the learning process with contextually grounded insights.

The operational dynamics of the \methodname are driven by the interaction between its various agents and the knowledge spaces,  as outlined in \cref{fig:clu}. Each component has a well-defined role in the learning and task-solving process, which can be understood through the interplay between task inputs, knowledge retrieval, prompt generation, and feedback-based refinement. These components work together to ensure the system's performance improves over time.

In the following sections, we will provide a detailed explanation of each critical component of the framework and then demonstrate the complete \textit{learning} and \textit{reasoning} mechanism.

\subsection{Task Execution through Knowledge and Prompt-Driven Agents} \label{subsec:task-agents}

\begin{figure*}[htb]
    \centering
    \includegraphics[width=\linewidth]{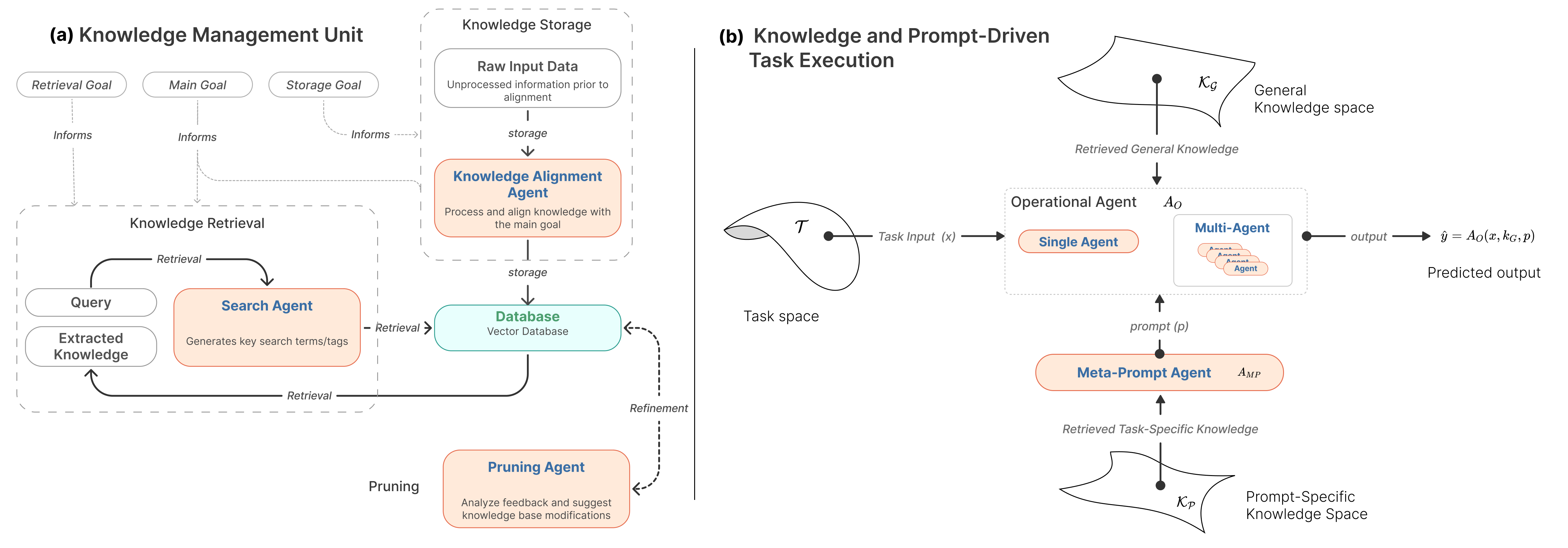}
    \caption{This figure shows the Knowledge Management Unit (KMU) and task execution flow, illustrating the interactions between knowledge storage, retrieval, and the operational agents for task-solving. In \textbf{(a)}, the KMU dynamically manages knowledge through three specialized agents: the Search Agent, which generates key search terms or tags for retrieving relevant knowledge; the Knowledge Alignment Agent, which processes raw input data \( \mathcal{I} \) and aligns it to the main goal before storing it in a vector database; and the Pruning Agent, which refines the stored knowledge based on feedback, ensuring that irrelevant or outdated information is pruned from the system. The two knowledge spaces, General Knowledge Space \( \mathcal{K_G} \) and Prompt-Specific Knowledge Space \( \mathcal{K_P} \), are dynamically informed and updated by the agents' operations. In \textbf{(b)}, the task space \( \mathcal{T} \) provides task inputs \( x \in \mathcal{T} \), which are processed by the Operational Agent \( A_O \). The Operational Agent can operate as a Single Agent or a Multi-Agent system, depending on the task complexity. The agent retrieves knowledge from the general \( \mathcal{K_G} \) and prompt-specific \( \mathcal{K_P} \) knowledge spaces. A Meta-Prompt Agent \( A_{MP} \) generates a prompt \( p \) based on the retrieved knowledge from \( \mathcal{K_P} \), guiding the Operational Agent in solving the task. The output \( \hat{y} \) is the result of the task, combining input \( x \), knowledge retrieval, and prompt generation. This combined figure showcases how feedback from task performance is used to iteratively refine the knowledge spaces, continuously improving the system's reasoning and execution capabilities.}
    \label{fig:combined_kmu}
\end{figure*}

To act effectively on a given task \( t \in \mathcal{T} \), the \methodnameshort\ relies on two crucial components: general knowledge about the broader task domain and a task-specific prompt. The general knowledge captures the high-level information required to perform the task in alignment with the global goal \( G \), while the task-specific prompt refines the action taken based on the nuances and specific requirements of the task at hand.

The general knowledge \( k_G \in \mathcal{K_G} \) is retrieved from the GKS using the retrieval function \( R_G(\mathcal{K_G}, t) \). This function extracts knowledge relevant to the overall goal and the task at hand, ensuring that the operational agent has access to domain-level information that informs its decision-making. In tandem, the task-specific prompt \( p \in \mathcal{P} \) is generated by the Meta-Prompt Agent \( A_{MP} \) using task-specific knowledge retrieved from \( \mathcal{K_P} \). The prompt \( p \) provides finer-grained instructions that allow the agent to adapt its actions to the specific task, incorporating any contextual details needed for accurate task execution.

The retrieval of both general knowledge and task-specific prompts, akin to retrieval-augmented generation (RAG) techniques with LLMs\cite{fan2024survey}, is formalized as:
\begin{equation}
    k_G = R_G(\mathcal{K_G}, t), \quad p = A_{MP}(t, R_P(\mathcal{K_P}, t))
\end{equation}

where \( R_P(\mathcal{K_P}, t) \) retrieves the task-specific information \( \mathcal{K_P} \), which the Meta-Prompt Agent uses to generate the prompt \( p \). Together, these two inputs provide the operational context in which the CLU operates, ensuring that both domain-level knowledge and task-specific insights are incorporated into the decision-making process.

The \textit{Operational Agent} \( A_O \) acts on the task by taking those above two critical inputs—general knowledge and the task-specific prompt—along with the task input \( x \in X_t \), to produce an output \( y \in Y \). The operational agent is the core component that processes the task by leveraging both broad and specific information. This agent can be designed as a singular or a multi-agent that acts as a single system when advanced reasoning is desired. For instance, where advanced reasoning is desirable, the operational agent might use multi-thought frameworks such as Chain-of-Thought (CoT)\cite{wei2022chain}, Tree-of-Thought (ToT)\cite{yao2024tree}, Everything of Thought \cite{ding2024thoughtsdefyinglawpenrose} or Iteration-of-Thought (IoT)\cite{radha2024iteration} to dynamically explore multiple decision paths or hypotheses before producing a final output. The flexibility of this agent allows the CLU  to scale from simple tasks to complex ones that require deep reasoning or sequential decision-making on top of the retrieved information.

Formally, the operational agent produces the output as:
\begin{equation}
    y = A_O(x, k_G, p) 
\end{equation}\label{eq:QA-output}

where \( k_G \) is the general knowledge retrieved from \( \mathcal{K_G} \), and \( p \) is the prompt generated from the task-specific knowledge \( k_P \in \mathcal{K_P} \) by Meta-Prompt agent. The \textit{Operational Agent} processes these inputs and generates the corresponding output \( y \). During the \textit{learning phase}, these outputs are then evaluated through the feedback mechanism, which will be further discussed in \cref{sec:OpDynamics}. A high level summary of  entire process is illustrated in \cref{fig:combined_kmu}(b).

As the system processes more tasks, the feedback mechanism provides essential information that helps the operational agent refine its decision-making process. This feedback allows the agent to learn from past iterations, progressively improving its performance. The iterative nature of this process ensures that the output \( y \) converges toward the expected result over time, either through supervised feedback in cases where labels are available or by aligning with the overarching goal \( G \) in unsupervised tasks. This feedback-driven refinement is crucial for enabling the \methodnameshort\ to handle a wide range of tasks and continuously update its knowledge spaces, leading to better performance over time. The combination of general knowledge and task-specific prompts provides a balanced approach to task execution, with the operational agent adapting its behavior based on the feedback loop to better align with both the task-specific and global goals.

It is important to emphasize that during the \textit{reasoning phase}, this component of the system—comprising the operational agent and the knowledge retrieval/prompt generation—serves as the final output-producing mechanism. In this phase, the general knowledge \( k_G \), task-specific prompt \( p \), and task input \( x \) are combined to generate the output \( y \), and no further steps are required. This ensures that the CLU can respond efficiently to new tasks in real-time. In contrast, during the \textit{learning phase}, this output is further evaluated through feedback mechanisms, which subsequently update the knowledge spaces \( \mathcal{K_G} \) and \( \mathcal{K_P} \), essentially modifying the knowledge spaces and tuning them toward the global goal \( G \). This distinction between \textit{learning} and \textit{reasoning} highlights the adaptability of the \methodnameshort\, where \textit{learning} improves performance over time while \textit{reasoning} allows for immediate output generation based on the current state of the knowledge spaces. Further discussion on these processes is given in \cref{sec:training-inference}.

\subsection{Knowledge Management Unit: Goal-Aligned Knowledge Transformation and Retrieval}\label{sec:KMU}
The use of external 'scratchpads' for intermediate computations has been shown to significantly improve the ability of language models to perform multi-step reasoning \cite{nye2021show}. In a similar vein, our knowledge spaces act as dynamic scratchpads, providing intermediate reasoning capabilities with the added advantage of continual refinement and long-term retention. The \textit{Knowledge Management Unit} (KMU), as a central innovation within our framework, orchestrates this dynamic knowledge storage and retrieval process through goal-oriented transformations, thereby elevating the utility of these knowledge spaces beyond static intermediates. Rather than serving as a traditional static repository, the KMU dynamically aligns stored information with overarching objectives, including the \textit{main goal}, the \textit{retrieval goal}, and the \textit{storage goal}. This alignment ensures that knowledge within the KMU is not only relevant but actively processed to serve the specific needs of both current and future tasks. Importantly, the KMU’s design is general enough to be applied outside of the CLU framework for tasks that require dynamic retrieval-augmented systems or continuous learning environments. As depicted in \cref{fig:combined_kmu}(a), the KMU uses specialized agents to handle data transformation, retrieval, and pruning, ensuring that the system remains efficient and relevant.

When new data is introduced into the system, it is not immediately stored as raw input. Instead, the \textit{Knowledge Alignment Agent} plays a crucial role in processing and aligning the data with the defined goals. This agent transforms the raw input into a more compact and relevant format that is optimized for future retrieval. The process includes extracting relevant information, adding tags aligned with the system’s main goal, and modifying the data as needed to ensure it fits the requirements of the storage goal. Formally, this transformation can be expressed as:
\begin{equation}
x_{\text{aligned}} = T(x, G_m, G_s)
\end{equation}
where \( T(\cdot) \) represents the transformation function applied by the Knowledge Alignment Agent, \( G_m \),   and \( G_s \) represent the main, and storage goals, respectively, and \( x_{\text{aligned}} \) is the aligned knowledge ready for storage. This knowledge is then stored in the respective knowledge spaces, either the GKS (\( \mathcal{K_G} \)) or the PKS (\( \mathcal{K_P} \)), depending on the nature of the information.

The KMU operates by first accepting raw input data. However, unlike conventional knowledge bases, it does not simply store this data as is. Instead, the \textit{Knowledge Alignment Agent} processes the data according to the main goal and the storage goal, aligning the raw input to ensure that it fits within the broader purpose of the system. This transformation is guided by the objectives set for the system, ensuring that knowledge is not only stored but refined and optimized for future retrieval. Mathematically, the KMU can be viewed as a function:
\begin{equation}
KMU = f(G_m, G_r, G_s, x) \rightarrow \mathcal{K_G}, \mathcal{K_P}
\end{equation}
where \( G_m \), \( G_r \), and \( G_s \) represent the main, retrieval, and storage goals, respectively, and \( x \) is the raw input data. The resulting output is the processed and aligned knowledge stored in the \textit{GKS} (\( \mathcal{K_G} \)) or the \textit{PKS} (\( \mathcal{K_P} \)), depending on the type of information.

In the \methodnameshort\ framework, we utilize two distinct KMUs. One is used to store broad, domain-specific knowledge that can generalize across multiple tasks, known as the GKS (\( \mathcal{K_G} \)), while the other is specifically tuned for task-specific insights and serves as the PKS (\( \mathcal{K_P} \)). The GKS acts as a foundation for reasoning and can be applied broadly across tasks. It retrieves domain-level information directly through the retrieval function:
\begin{equation}
k_G = R_G(\mathcal{K_G}, G_r, t)
\end{equation}
where \( R_G(\cdot) \) extracts the relevant general knowledge based on the retrieval goal \( G_r \) and the task \( t \). The PKS, on the other hand, stores information tailored to task-specific prompts and adjustments. It is retrieved using:
\begin{equation}
k_P = R_P(\mathcal{K_P}, G_r, t)
\end{equation}
where \( R_P(\cdot) \) retrieves task-specific knowledge that informs prompt generation and behavior adjustment.

This dual-knowledge setup allows the KMU to support both generalized reasoning and task-specific fine-tuning, creating a balanced and adaptable knowledge management system that can efficiently handle a wide range of tasks.


To maintain efficiency, KMU employs a pruning mechanism that uses feedback from task execution, analyzed by the \textit{Pruning Agent}, to refine the knowledge base. This ensures that outdated or irrelevant information is systematically removed, preventing the accumulation of incorrect knowledge, which is especially prevalent during early learning phases akin to random initialization in deep neural networks (DNN). Unlike traditional DNNs, where weight adjustments through backpropagation enhance performance by discarding ineffective parameters, pruning in KMU targets the knowledge level, allowing for continuous realignment of knowledge with evolving system objectives. Formally, pruning is expressed as:
\begin{equation}
\mathcal{K_G} = P_G(\mathcal{K_G}, \mathcal{F}^n), \quad \mathcal{K_P} = P_P(\mathcal{K_P}, \mathcal{F}^n)
\end{equation}
where \( P_G(\cdot) \) and \( P_P(\cdot) \) represent the pruning functions for the general and PKSs, respectively, based on feedback \( \mathcal{F}^n \) accumulated over \( n \) iterations.

The importance of pruning in this context cannot be understated. As the system evolves and processes more tasks, outdated or irrelevant knowledge could otherwise accumulate, slowing down retrieval processes and potentially leading to incorrect inferences. By integrating feedback-based pruning, the KMU ensures that only the most relevant and accurate knowledge remains in the system, thereby improving efficiency and task performance.

The versatility of the KMU extends beyond its use in the \methodname framework. Its modular design allows it to be applied to a wide range of AI-driven applications, particularly those that require efficient knowledge retrieval and dynamic storage management. In retrieval-augmented generation systems, for example, the KMU’s ability to process and align knowledge based on specific goals would enable more efficient and targeted knowledge retrieval, while its pruning mechanism would ensure that only the most relevant information is maintained over time. Similarly, in continuous learning environments, the KMU could serve as a dynamic memory system that evolves alongside the system, ensuring that knowledge is consistently updated and aligned with evolving goals. Exact implementation details of KMU is given in \cref{alg:kmu_operations}.

\begin{algorithm}[!htb]
\caption{Knowledge Management Unit (KMU) Operations}
\label{alg:kmu_operations}
\begin{algorithmic}
\Require Main goal $G_m$, Retrieval goal $G_r$, Storage goal $G_s$
\Ensure Optimized knowledge spaces $\mathcal{K_G}$, $\mathcal{K_P}$

\Function{SaveKnowledge}{$x$}
    \State $x_{\text{aligned}} \gets T(x, G_m, G_s)$ \Comment{Align knowledge based on main and storage goals}
    \State $\text{id} \gets \text{GenerateUniqueID}()$ \Comment{Assign unique ID to aligned knowledge}
    \State $e \gets \text{ComputeEmbedding}(x_{\text{aligned}})$ \Comment{Compute embedding of aligned data}
    \State $\text{StoreInDatabase}(x_{\text{aligned}}, e, \text{id})$ \Comment{Store aligned data and its embedding}
    \State \Return id
\EndFunction

\Function{RetrieveKnowledge}{$q, n$}
    \State $t \gets A_{\text{S}}(q, G_m, G_r)$ \Comment{Generate search terms based on query, main, and retrieval goals}
    \State $q_{\text{combined}} \gets \text{CombineTerms}(t)$ \Comment{Combine search terms for effective query}
    \State $e_q \gets \text{ComputeEmbedding}(q_{\text{combined}})$ \Comment{Compute embedding of combined query}
    \State $\text{results} \gets \text{QueryDatabase}(e_q, n)$ \Comment{Retrieve top $n$ results from the database}
    \State \Return results
\EndFunction

\Function{PruneKnowledge}{$f, \text{ids}$}
    \State $x_{\text{existing}} \gets \text{GetEntries}(\text{ids})$ \Comment{Retrieve existing entries based on IDs}
    \State $x_{\text{new}} \gets A_{\text{P}}(f, x_{\text{existing}}, G_m, G_s)$ \Comment{Prune and update knowledge based on feedback}
    \State $\text{DeleteEntries}(\text{ids})$ \Comment{Delete outdated or irrelevant entries from the database}
    \State $\text{new\_ids} \gets \emptyset$
    \For{each $x_i$ in $x_{\text{new}}$}
        \State $\text{id}_i \gets \text{SaveKnowledge}(x_i)$ \Comment{Save updated knowledge and assign new IDs}
        \State $\text{new\_ids} \gets \text{new\_ids} \cup \{\text{id}_i\}$
    \EndFor
    \State \Return new\_ids
\EndFunction

\end{algorithmic}
\end{algorithm}

\subsection{Feedback Mechanism and Knowledge Update}\label{sec:feedback}

The core strength of the CLU framework lies in its ability to adapt and improve continuously through feedback. Rather than relying on static knowledge or manually curated updates, CLU dynamically evolves by incorporating feedback signals from task performance. This feedback not only informs the system about the correctness of its outputs but also provides actionable insights that guide the refinement of its internal knowledge stores. The flow of this feedback process, from task execution to knowledge storage, is illustrated in \cref{fig:clu}. In \methodnameshort\, the feedback mechanism starts by analyzing the output of the task generated by the Operational Agent ($A_O$). Depending on whether the task is supervised or unsupervised, the system compares the output $y$ to an expected output $y^*$ (in supervised cases) or evaluates performance metrics aligned with the main goal $G$ (in unsupervised cases). This comparison is handled by the \textit{Response Comparison Agent}, which determines whether the task was successfully completed. Formally, the comparison can be expressed as:

\begin{equation}
    c = A_{GF}(y, y^*, t) \label{eq:comparison}
\end{equation}

where $c$ represents the feedback comparison result, $y$ is the generated output, and $y^*$ is the expected result (if available).
\subsubsection{Feedback Agents}
The \textit{Response Comparison Agent} routes the feedback  $c$ (from \cref{eq:comparison}) to one of two agents: the \textit{Positive Feedback Agent} ($A_{PF}$) or the \textit{Negative Feedback Agent} ($A_{NF}$). These agents serve distinct roles in reinforcing successful patterns and identifying areas for improvement, respectively.

\textbf{Positive Feedback Agent ($A_{PF}$)}:
When the task performance aligns with expectations, the Positive Feedback Agent is triggered. Its purpose is to amplify the patterns that led to success, reinforcing the system's knowledge and prompting it to prioritize similar strategies in the future. This reinforcement helps the system store successful task patterns in the KMS.

\textbf{Negative Feedback Agent ($A_{NF}$)}:
Conversely, if the task output deviates from expectations, the Negative Feedback Agent steps in to identify errors. It analyzes where the system went wrong and refines the knowledge and task-specific strategies stored in the KMS to correct the issue. This ensures that the system learns from its mistakes and adapts accordingly.

Both of these agents operate based on the feedback signal $f$, which is determined by the comparison result $c$:

\begin{equation}
f =
\begin{cases}
A_{PF}(c, y, t), & \text{if } c \in \mathcal{C}_{\text{success}} \\
A_{NF}(c, y, t), & \text{if } c \in \mathcal{C}_{\text{failure}}
\end{cases}.
\label{eq:feedback_signal}
\end{equation}\label{eq:SFT-compare}

Here, $c$ determines whether the system receives positive or negative feedback, depending on how the output is evaluated. In the unsupervised scenario, where no ground-truth output $y^*$ is available, a \textit{General Feedback Agent} ($A_{GF}$) is employed to evaluate how well the generated output $y$ aligns with the overarching goal $G$, providing a qualitative feedback signal for knowledge refinement.

\subsubsection{Knowledge Extraction and Insight}

Once feedback is processed, the \textit{Knowledge Insight Agent} ($A_{KI}$) plays a pivotal role in extracting actionable information from the feedback and routing it to the relevant knowledge spaces. The agent synthesizes new knowledge from the feedback signal $f$ and integrates this knowledge into the system’s KMS. The extracted knowledge, denoted as $k_{\text{new}}$, is a direct result of analyzing both successful and unsuccessful task completions:

\begin{equation}
    k_{\text{new}} = A_{KI}(f, t, y) \label{eq:new_knowledge}
\end{equation}

The new knowledge $k_{\text{new}}$ is then added to either the GKS ($\mathcal{K_G}$) or the PKS ($\mathcal{K_P}$), depending on its nature:

\begin{equation}
    \mathcal{K_G} = U_G(\mathcal{K_G}, k_{\text{new}}), \quad \mathcal{K_P} = U_P(\mathcal{K_P}, k_{\text{new}}) \label{eq:update_knowledge}
\end{equation}

Here, $U_G$ and $U_P$ represent the update functions for the general and PKSs, respectively. This process ensures that the system continuously refines its knowledge based on the feedback it receives.

\subsubsection{Knowledge Pruning Mechanism}\label{sec:pruning}

As \methodnameshort\ processes more tasks, irrelevant or outdated knowledge may accumulate in the knowledge spaces. To prevent this, the system employs a \textit{pruning mechanism}, which ensures that only relevant knowledge is retained. The pruning process is triggered periodically, after a set number of iterations ($n$), and is based on a history of feedback signals $\mathcal{F}^n$. This pruning is crucial for maintaining the efficiency and accuracy of the knowledge retrieval process.

The pruning mechanism is expressed as:

\begin{equation}
    \mathcal{K_G} = P_G(\mathcal{K_G}, \mathcal{F}^n), \quad \mathcal{K_P} = P_P(\mathcal{K_P}, \mathcal{F}^n) \label{eq:pruning}
\end{equation}

Where $P_G$ and $P_P$ represent the pruning functions for the general and PKSs. The feedback history $\mathcal{F}^n$ ensures that knowledge is evaluated periodically for relevance and utility.

In summary, the feedback mechanism in \methodnameshort\ is critical to its adaptability and long-term learning. By processing task performance and routing feedback through specialized agents, the system is able to reinforce successful patterns and correct errors efficiently. The extracted knowledge from the feedback is stored in dynamic knowledge spaces, ensuring that the system remains aligned with the overarching goal $G$. Furthermore, the periodic pruning of irrelevant knowledge ensures that the system remains efficient and focused as it evolves over time.

\begin{algorithm*}[!htb]
\caption{Meta-Learning Unit (MLU) Adaptation and Reasoning Process}
\label{alg:mlu_process}
\begin{algorithmic}[1]
\Require Task set $\mathcal{T}$, Quality metric $Q$, Operational Agent $A_O$, Initial KMS $\mathcal{K}_0$
\Ensure Optimized KMS $\mathcal{K}^{*}$
\State $\mathcal{F}_{\text{history}} \gets \emptyset$ 
\For{each task $t_i \in \mathcal{T}$}
    \State $x_i \gets \text{InputData}(t_i)$ 
    \State $k_{G_i} \gets R_G(\mathcal{K}_G, t_i)$ \Comment{Retrieve general knowledge}
    \State $k_{P_i} \gets R_P(\mathcal{K}_P, t_i)$ \Comment{Retrieve prompt knowledge}
    \State $p_i \gets A_{MP}(t_i, k_{P_i})$ \Comment{Generate prompt}
    \State $y_i \gets A_O(x_i, k_{G_i}, p_i)$ \Comment{Execute task}
    \If{TrainingMode}
        \State $y^*_i \gets \text{ExpectedOutput}(t_i)$ \Comment{Get ground truth}
        \State $c_i \gets A_{GF}(y_i, y^*_i, t_i)$ \Comment{General feedback}
        \If{$c_i \in \mathcal{C}_{\text{success}}$}
            \State $f_i \gets A_{PF}(c_i, y_i, t_i)$ \Comment{Positive Feedback}
        \Else
            \State $f_i \gets A_{NF}(c_i, y_i, t_i)$ \Comment{Negative Feedback}
        \EndIf
        \State $\mathcal{K}_{\text{new}} \gets A_{KI}(f_i, t_i, y_i)$ \Comment{Extract knowledge}
        \ForAll{$k_{\text{new}} \in \mathcal{K}_{\text{new}}$}
            \State $\mathcal{K}_G \gets U_G(\mathcal{K}_G, k_{\text{new}})$ \Comment{Update general KMS}
            \State $\mathcal{K}_P \gets U_P(\mathcal{K}_P, k_{\text{new}})$ \Comment{Update prompt KMS}
        \EndFor
        \State $\mathcal{F}_{\text{history}} \gets \mathcal{F}_{\text{history}} \cup \{f_i\}$ 
        \If{$|\mathcal{F}_{\text{history}}| \geq n$} \Comment{Prune periodically}
            \State $\mathcal{K}_G \gets P_G(\mathcal{K}_G, \mathcal{F}_{\text{history}})$ 
            \State $\mathcal{K}_P \gets P_P(\mathcal{K}_P, \mathcal{F}_{\text{history}})$ 
            \State $\mathcal{F}_{\text{history}} \gets \emptyset$ 
        \EndIf
    \EndIf
\EndFor
\State $\mathcal{K}^{*} \gets \underset{\mathcal{K}}{\operatorname{argmax}} \ \mathbb{E}_{\substack{t \sim \mathcal{T} \\ x \sim X_t}} \left[ Q(A_O(x, R_G(\mathcal{K}_G, t), A_{MP}(t, R_P(\mathcal{K}_P, t)))) \right]$
\State \Return $\mathcal{K}^*$
\end{algorithmic}

\caption{The full \methodname algorithm outlining both the training and inference processes. During training, the system retrieves knowledge, generates prompts, executes tasks, and incorporates feedback to update the Knowledge Management System (KMS). Inference mode skips the feedback incorporation, focusing on efficient task execution using the current knowledge. Periodic pruning ensures that the KMS remains optimized over time.}
\end{algorithm*}

\subsection{Operational Dynamics: Learninng and reasoning}\label{sec:OpDynamics}

The operational dynamics of the Composite Learning Unit (CLU) encompass two distinct but closely related processes: the \textit{learning phase} and the \textit{reasoning phase}. These phases are not separate in a strict sense; instead, reasoning is an integral part of learning, making the system inherently adaptable. During the learning phase, the CLU refines its knowledge through continuous interaction, leveraging feedback mechanisms to update its knowledge spaces. This allows for ongoing improvement, aligning knowledge to better meet the overarching goal across diverse tasks. In contrast, the reasoning phase focuses on efficiently utilizing the current state of knowledge to perform task-specific actions, generating outputs without modifying internal representations. As shown in \cref{fig:clu}, reasoning is a subset of the broader learning process, meaning that the CLU can dynamically transition between reasoning and learning as needed. This capability enables the system to perform real-time reasoning while maintaining the option to incorporate learning whenever a decrease in performance or a knowledge gap is detected. The flexibility to invoke learning during reasoning is highly advantageous, allowing the system to adjust to changes dynamically, albeit at the cost of increased computational requirements. Importantly, this adaptive approach ensures that the CLU remains both responsive and continuously evolving, tailored to meet the specific demands of its operating environment.

The \textit{reasoning phase} of the Composite Learning Unit (CLU) serves as the foundation for task execution, providing a streamlined approach for generating outputs based on the current knowledge state. During this phase, the system retrieves relevant knowledge from the GKS (\(\mathcal{K_G}\)) and task-specific prompts from the PKS (\(\mathcal{K_P}\)). The operational agent \( A_O \) utilizes this knowledge to produce an output \( y \), as described in \cref{eq:QA-output}, focusing solely on effective task execution without altering its internal knowledge base. This \textit{reasoning} process allows the CLU to respond swiftly to new tasks, leveraging the established knowledge without engaging in feedback-based refinement. Consequently, \textit{reasoning} is computationally efficient, yielding outputs without the complexity of knowledge updates.

Building upon the reasoning process, the \textit{learning phase} introduces an iterative, feedback-driven approach aimed at refining the system’s knowledge spaces. During this phase, the CLU executes tasks while evaluating its output against expected results or overarching goals, as formalized in \cref{eq:SFT-compare}. Feedback generated from this comparison is routed through either the Positive Feedback Agent ($A_{PF}$) or the Negative Feedback Agent ($A_{NF}$), facilitating reinforcement of successful strategies or corrective measures for erroneous outputs. The knowledge insights ($k_{\text{new}}$) are subsequently integrated into both $\mathcal{K_G}$ and $\mathcal{K_P}$, enabling the system to improve its task performance iteratively. The overall objective, expressed in \cref{eq:maximize_exp}, is to optimize knowledge representations within the Knowledge Management System (KMS), enhancing the efficacy of task-solving capabilities over time. Importantly, the distinction between reasoning and learning phases is largely virtual, arising primarily from current computational cost and speed limitations. As these limitations diminish, the CLU could operate seamlessly in both phases, continually refining knowledge while performing tasks, akin to continual learning during inference. This dual focus on immediate execution and ongoing improvement ensures that the CLU evolves with increasing sophistication across diverse tasks, as detailed in \cref{alg:mlu_process}.

\section{Discussion}\label{sec:training-inference}

The CLU's training dynamics introduce a novel approach to learning, emphasizing continuous knowledge refinement through feedback-driven adaptation rather than parameter tuning via weight updates. In the \textit{learning phase}, CLU integrates knowledge from both general and task-specific contexts, adjusting these dynamically to enhance task performance over time. Unlike traditional deep neural networks (DNNs) that rely on backpropagation, where $\Delta \theta = -\eta \nabla_\theta \mathcal{L}(\theta)$ \cite{rumelhart1986learning}, the CLU refines its knowledge through iterative reasoning. Notably, while CLU's approach diverges from traditional DNNs, the underlying large language models (LLMs) used in CLU are still trained using conventional backpropagation methods. Here, we leverage these LLMs as a base and "dress" them to make them capable learners, enabling CLU to adapt to new tasks through continuous refinement. Instead of static parameter ($\theta$) optimization, the system continuously adjusts its GKS ($\mathcal{K_G}$) and PKS ($\mathcal{K_P}$) through feedback, ensuring that both generalizable and task-specific knowledge are progressively enhanced (\cref{eq
}). The \textit{reasoning phase} operates differently, focusing solely on generating outputs from the current state of knowledge without further refinement, thereby facilitating efficient and direct response to new tasks (\cref{sec:training-inference}). 

At the core of the CLU framework are base reasoners, specifically in current work, large language models (LLMs), which provide foundational reasoning capabilities \cite{brown2020language,openai2023gpt4}. The learning framework revolves around these reasoning agents, where the sophistication of the underlying reasoners directly influences the learning efficacy and adaptability of the entire system. The more advanced the base reasoning capabilities, the more effective the learning and task performance of the unit. While this work leverages LLMs for reasoning, the conceptual framework is inherently general and can be extended to any future reasoning systems, including multimodal models involving language, speech, vision, or even symbolic reasoning models\cite{trinh2024solving} that utilize logic-based inference mechanisms such as deduction, induction, and abduction. This adaptability makes the CLU a highly flexible learning architecture, capable of leveraging different reasoning paradigms to evolve through feedback. By distinguishing between the learning and reasoning phases, the CLU allows for continuous knowledge-driven adaptation, ensuring that it remains responsive and progressively improves based on task-specific interactions. In the following sections, we delve deeper into how inference operates as an active reasoning process, detailing its role in continuous adaptation, computational efficiency, and dynamic knowledge management across diverse use cases.

\paragraph{Reasoning as Continual Learning:} Unlike traditional deep learning models, where inference is merely a forward pass through fixed parameters, reasoning in CLU incorporates continual learning by directly integrating feedback into its operational cycle. Each reasoning step not only generates an output but also contributes to refining knowledge spaces (\(\mathcal{K_G}\) and \(\mathcal{K_P}\)), allowing real-time adaptation without weight updates (\cref{sec:OpDynamics}). This makes reasoning in CLU inherently adaptive, unlike static models that require costly retraining. By dynamically refining knowledge based on immediate feedback, CLU blurs the boundary between training and inference, enabling a continuous improvement process well-suited to evolving and varied task environments.

\paragraph{Computational Efficiency and Adaptability:}  CLUs address main limitation of current LLMs which requires frequent computationally demanding and expensive retraining or fine tuning to keep up with evolving knowledge sources and tasks spaces. Retraining such models to incorporate new data or fine-tune for evolving tasks demands significant computational resources, particularly given their  ever increasing model parameters currently reaching trillions\cite{zhang2024scaling}, often requiring extensive compute infrastructure that limits adaptability in real-time or dynamic environments. In contrast, CLU employs a feedback-driven learning approach that circumvents parameter updates by continuously refining its knowledge spaces. Unlike deep neural networks that depend on backpropagation for weight adjustment, CLU leverages lightweight inference steps (of underlying reasoners, which is LLMs in our case) to modify its internal knowledge directly, enabling computational  efficiency and real-time adaptation. As detailed in \cref{sec:feedback}, this feedback mechanism supports rapid integration of new information without the resource-intensive retraining cycles characteristic of traditional models. This design shares conceptual similarities with recent retrieval-augmented generation (RAG) models \cite{lewis2020retrieval}, which optimize responses using externalized memory, but CLU extends beyond RAG by incorporating continual feedback to refine both general and prompt-specific knowledge, facilitating real-time adaptability across tasks. Consequently, CLU dynamically adjusts to evolving requirements without the computational cost of fine-tuning, bridging the gap between inference and learning efficiently across diverse environments.

\paragraph{Continuous Feedback-Driven Adaptation:} CLU’s feedback mechanism is philosophically akin to backpropagation, in that both use feedback to improve performance over time. However, instead of adjusting model weights, CLU refines its knowledge spaces. Positive feedback reinforces successful knowledge usage, while negative feedback triggers refinement in representations or prompt strategies (\cref{sec:feedback}). In supervised tasks, feedback involves comparing outputs against ground truth (\cref{eq:SFT-compare}), while in unsupervised tasks, it assesses goal alignment. This approach shares similarities with reinforcement learning, where feedback acts like a reward signal that guides the refinement of knowledge spaces rather than updating parameters. Unlike traditional models, which require backpropagation or retraining for adaptation, CLU's feedback loop directly refines general and prompt-specific knowledge in real time, allowing for continuous adaptation. This fusion of RL-like feedback with continual learning principles enables CLU to improve its task performance across varied environments, bridging the gap between inference and learning without the computational expense of weight updates.

\paragraph{Dynamic Knowledge Management and Real-Time Adaptation}
Traditional DNN models are often plagued by issues like catastrophic forgetting, where fine-tuning on new tasks leads to the degradation of performance on previously learned tasks \cite{french1999catastrophic}. In contrast, CLU's architecture is designed to retain general knowledge across tasks while allowing task-specific adaptation through its dual knowledge spaces. The GKS (\(\mathcal{K_G}\)) stores broad domain-level knowledge, while the PKS (\(\mathcal{K_P}\)) captures fine-grained task details. This separation enables CLU to adapt without overwriting previously learned knowledge, making it better suited for dynamic, evolving tasks. The feedback incorporation and subsequent knowledge refinement are formalized in \cref{alg:mlu_process}, where the system continuously updates its knowledge spaces based on feedback from task execution. This mechanism allows CLU to leverage feedback in a manner similar to reinforcement learning, continuously improving both \(\mathcal{K_G}\) and \(\mathcal{K_P}\) for better adaptability. Moreover, CLU's approach bears similarities to Memory Networks \cite{weston2015memorynetworks}, which also employ external memory to store and retrieve information relevant to complex queries. However, unlike Memory Networks that use static memory interaction primarily during inference, CLU employs a continuous feedback-driven process to dynamically update its knowledge spaces, allowing it to refine and prune its stored information in real time (\cref{sec:pruning}). This feedback and pruning mechanism ensures that CLU maintains a lean, relevant knowledge base, preventing the accumulation of redundant or outdated knowledge that often plagues traditional models. Consequently, CLU can adapt efficiently to new tasks without extensive retraining, ensuring robust performance in dynamic environments.

\paragraph{Applications Beyond Traditional Training}
CLU's flexible framework allows it to address problems beyond the capabilities of traditional deep neural networks, which often focus on static, predefined tasks and representation learning. Unlike these models that specialize in mapping inputs to outputs or extracting feature representations, CLU is inherently designed to reason continuously and adaptively about high-level latent information, even when confronted with dynamically evolving data. Through its goal-directed reasoning process, CLU can autonomously discover and refine complex patterns over time, extracting knowledge such as relationships and abstract attributes from unstructured data without requiring retraining. This adaptability is facilitated by its dual knowledge management system (\cref{subsec:general-arcitecture}), consisting of the GKS and the PKS. Together, these knowledge spaces enable CLU to effectively adapt its learning to the requirements of incoming data streams, supporting applications in meta-learning, real-time decision-making, and agent-driven simulations—particularly those that demand continuous adaptation and abstraction. By utilizing inference-based reasoning augmented by continuous feedback, CLU ensures that meaningful patterns are efficiently captured and retained, achieving an adaptive response where traditional DNNs would require computationally intensive retraining.

Unlike conventional deep learning systems that remain fixed after training, CLU exemplifies an adaptive learning architecture, capable of addressing evolving and complex tasks through continuous reasoning and feedback-driven knowledge refinement. Instead of passively fitting input-output pairs, CLU actively reasons about the latent relationships and high-level abstractions that emerge across diverse environments, dynamically refining both domain-specific and task-specific knowledge as required. This ability to engage in continual adaptation extends beyond the traditional paradigm of training and inference, allowing CLU to efficiently tackle new, unseen challenges by abstracting, reasoning, and adapting without retraining. Whether the problem involves supervised learning, unsupervised discovery, or abstract inference, CLU leverages its accumulated knowledge to solve complex problems more effectively, achieving a level of versatility and adaptability beyond that of traditional deep learning models.

To summarize, CLU is designed to adapt dynamically to a broad range of tasks by aligning its learning process with task-specific and abstract goals. Unlike traditional deep learning systems, which often rely on static parameter adjustments, CLU generalizes across both supervised and unsupervised settings by continuously retrieving and updating knowledge from its internal knowledge spaces. This adaptability allows CLU to refine its understanding of evolving task requirements without extensive retraining, making it particularly effective in environments that require real-time learning and adaptation. CLU's architecture separates domain-level and task-specific knowledge, which facilitates its capability to reason about complex relationships and draw abstract insights, even when confronted with unstructured or dynamic data. Whether learning relationships between data points in an unsupervised corpus or adapting recommendations based on user interactions, CLU refines its internal representations and retains the flexibility to respond to high-level queries about its accumulated knowledge.

Furthermore, CLU's modular architecture allows individual learning units to work independently or collaboratively within a broader composite system. Each unit can be assigned a unique goal, and collectively they can contribute to a larger objective, enhancing the overall system's ability to address multifaceted problems through specialized reasoning capabilities. This meta-level organization of CLUs enables an adaptive learning framework where specialized units continuously evolve to tackle distinct aspects of complex tasks, a capability that surpasses traditional neural networks or standalone reasoners—a direction we reserve for exploration in future work.

In the following section, we provide an empirical demonstration of CLU's adaptive reasoning framework in action, showcasing its ability to learn complex tasks and offering a practical glimpse into its application, where even traditional reasoners made from deep neural networks (in this specific case large language models), often fail.

\section{Illustrative Example of Adaptive Learning in Cryptographic Reasoning}\label{sec:results}
\setcounter{problem}{1}

The task of evaluating an intelligent system's adaptive reasoning capabilities requires a problem that extends beyond simple pattern recognition. We sought a problem that would not only test the system's ability to recognize intricate relationships but also demand continuous adaptation through iterative refinement. Such a problem should require reasoning at multiple levels—extracting rules, hypothesizing solutions, learning from errors, and refining strategies over time—mirroring the way intelligent agents operate in real-world scenarios. To this end, we designed the cryptographic problem described in \problemref{probdef}, which involves deducing an encoding rule from a limited set of examples. While theoretically straightforward, as we will demonstrate, the underlying base reasoners struggle to solve this problem effectively, making it a suitable challenge for assessing the capabilities of our proposed Composite Learning Unit.

As discussed in \cref{sec:introduction}, a system grounded in Constructivism, Active Inference, and Information Foraging is well-suited for this challenge. Constructivism supports iterative knowledge building, allowing CLU to refine its understanding of the encoding rule with each new example. Active Inference provides a framework for reducing uncertainty through hypothesis generation and testing, which is crucial for inferring complex patterns. By iteratively seeking and verifying rules, CLU actively adapts to new tasks. This problem aligns with CLU's core objectives, directly testing its capacity for dynamic adaptation, abstraction, and continual learning—qualities that traditional static models lack.

In the following evaluations CLU uses \code{GPT-4o-mini} model as the underlying base LLM model for reasoning. The evaluation involved generating a dataset of $n + 150$ sentences, where $n$ represents the number of shots in "n-shot learning," varying between 1 and 5. The first $n$ examples were used to establish the encoding rule, while the remaining 150 sentences were utilized for testing the performance. The dataset was constructed with diverse sentences to avoid any contextual dependencies, ensuring that the problem required genuine reasoning abilities rather than simple memorization. 

\begin{figure*}[tb]
    \centering
    \includegraphics[width=\linewidth]{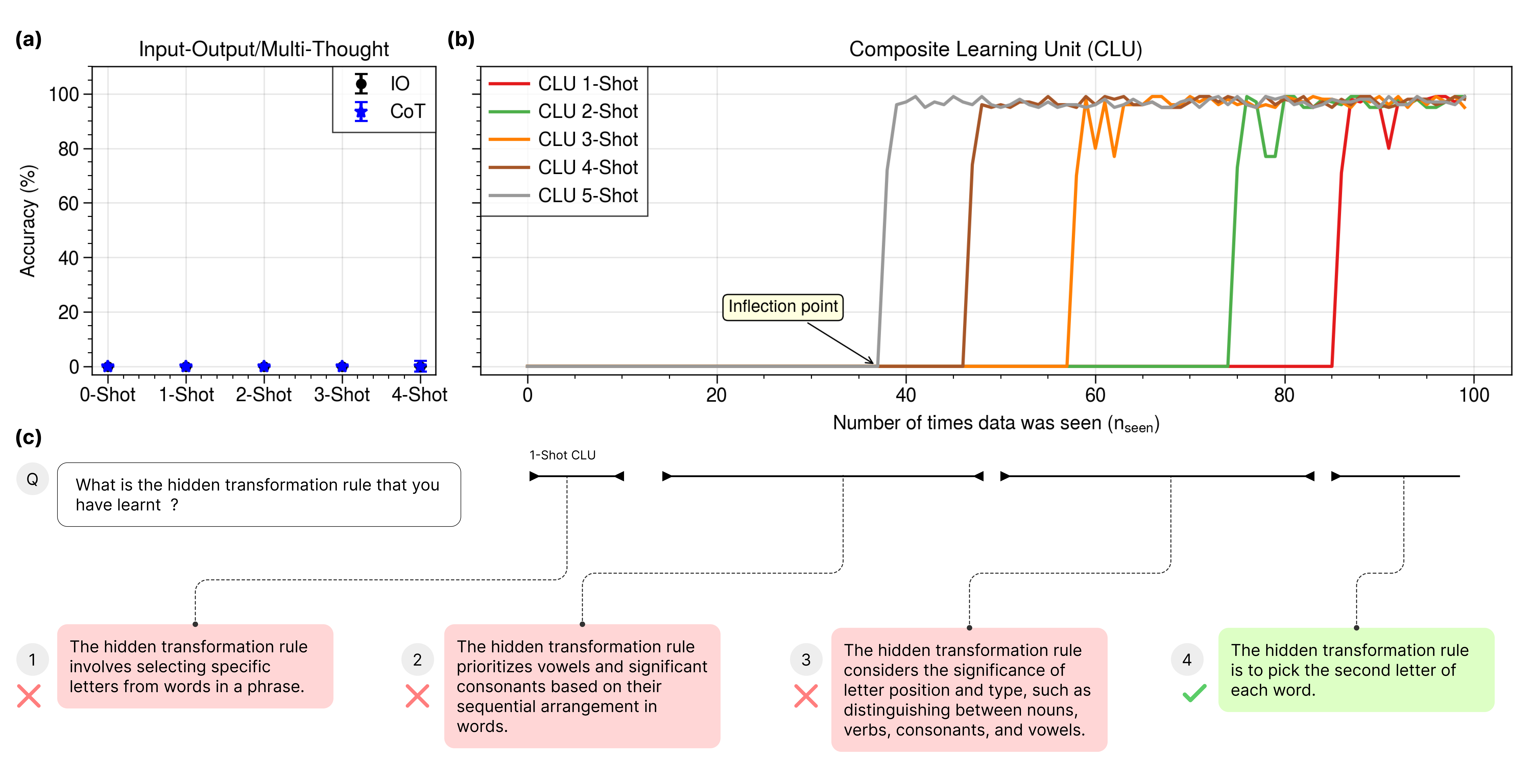}
    \caption{\textbf{(a)} \textit{Baseline Performance with IO/CoT:} The performance of the baseline Input-Output (IO) and Chain-of-Thought (CoT) methods is depicted here, showing $0\%$ accuracy for all shots, from 0-shot to 4-shot. This indicates the inability of the underlying GPT-4o-mini model to infer the correct transformation rule, even with increased examples and guided reasoning prompts. \textbf{(b)} \textit{Learning Dynamics of Composite Learning Unit (CLU):} The CLU's performance over multiple learning iterations is presented. Initially, the accuracy remains low, but after an inflection point, CLU's performance rapidly increases, eventually stabilizing near $100\%$. This improvement is driven by CLU's iterative refinement of its knowledge, facilitated by feedback mechanisms that guide the system to correctly infer the transformation rule, as detailed in \cref{sec:results}. \textbf{(c)} \textit{Evolution of the Transformation Rule Understanding:} The evolution of CLU's internal understanding is shown here, illustrating the progression of hypotheses across learning iterations for the 1-shot setting. Initially, CLU makes vague or incorrect guesses regarding the transformation rule, but as it iteratively incorporates feedback, it eventually converges on the correct rule of selecting the second letter from each word.
    }
    \label{fig:results}
\end{figure*}

The results of the baseline model's Input-Output (IO) and Chain-of-Thought (CoT) enhanced reasoning are summarized in \cref{fig:results}. In the IO case, the \code{GPT-4o-mini} model was directly prompted with the new sentence after being given $n$ examples, resulting in consistently $0\%$ accuracy across all shots. In the case of CoT evaluation, where the base \code{GPT-4o-mini} model was prompted step-by-step to enhance its reasoning capabilities, it failed to go beyond the base model's performance of $0\%$  accuracy, indicating that even guided reasoning with the base model was inadequate for inferring the solution. This suggests that the inherent complexity of the encoding rule, despite being simple in theory, was beyond the capabilities of the baseline reasoners.
\begin{tcolorbox}[colback=gray!10, colframe=black, title={\theproblem}, label={probdef}]
    Given a set of \(n\)-shot examples, where each example consists of a sentence and its encoded output, the goal is to derive a specific encoding pattern for a new sentence. The encoding rule is defined as selecting the \(i\)-th letter from each word in the given sentence, excluding words with fewer than \(i\) letters. Formally, for a given sentence \(S = (w_1, w_2, \dots, w_k)\), the output is \(E = (c_1, c_2, \dots, c_m)\), where each \(c_j\) corresponds to the \(n\)-th character of the word \(w_j\) if \(\text{len}(w_j) \geq n\). The task is to infer the encoding pattern from \(n\)-shot examples and then apply this pattern to new input sentences to produce the correct encoded output.\\

    \textbf{Example for 3-shot query with $i = 2$}

    Given:
    \begin{itemize}
        \item "Neural networks transform data efficiently" \(\rightarrow\) "eeraf"
        \item "Artificial intelligence automates decisions" \(\rightarrow\) "rnue"
        \item "Amazing Large language models" \(\rightarrow\) "maao"
    \end{itemize}
    Query:
    \begin{itemize}
        \item What is "Gradient descent optimizes loss functions" \(\rightarrow\) ?
    \end{itemize}
\end{tcolorbox}
The CLU framework was subsequently evaluated by repeatedly training on the same $n$-shot examples, enabling the system to refine its understanding of the underlying transformation rule iteratively. For each $n$-shot setup, CLU was trained serially over a set number of iterations, denoted as $n_{seen}$, where the main goal $G_m$ was specified as: \code{Find the hidden rule and learn the transformation logic}. During training, pruning was triggered after every five iterations to remove or adjust irrelevant knowledge, allowing CLU to maintain a concise and relevant knowledge base. During testing, CLU was set to operate in the \textit{reasoning} state, where examples were used to evaluate performance without contributing to feedback or knowledge accumulation. The accuracy performance for different $n$-shot settings over multiple learning iterations is presented in  \cref{fig:results}(b).

Unlike traditional training methods that risk overfitting through repeated exposure to identical data, CLU’s learning process is inherently goal-driven, focusing on understanding and abstracting the transformation logic as per \(G_m\). While defining \(G_m\) as ``learning input-output pairs'' would lead to memorization, setting \(G_m\) as ``learning the transformation logic'' allows CLU to shift its focus to acquiring abstract, underlying rules. Through the iterative refinement of its knowledge spaces (\cref{subsec:general-arcitecture}), CLU adapts its learning to effectively generalize the encoding process, ensuring that repeated exposure fosters deeper comprehension rather than overfitting. In the initial iterations, CLU's performance was comparable to traditional inference-based methods, as reflected by the $0\%$ accuracy seen in the early segment of the plots in \cref{fig:results}(b). This was anticipated since CLU begins with an empty knowledge base, functioning similarly to a basic Input-Output mechanism, with the operational agent generating prompts purely based on the main goal $G_m$. The accuracy remains at $0\%$ across several initial iterations, as the system lacks any accumulated knowledge to discern the hidden transformation pattern effectively. During this stage, CLU essentially relies on trial and error, leading to repeated failures and the subsequent accumulation of negative feedback, which ultimately lays the groundwork for future knowledge refinement.

A critical observation emerges around the inflection point shown in \cref{fig:results}(b), where accuracy rapidly escalates towards near-perfect performance. This inflection point signifies the moment when CLU, after accumulating and synthesizing multiple rounds of feedback, begins to recognize and apply key aspects of the underlying transformation rule. The system's persistence through negative feedback during the early learning stages leads it to explore various hypotheses and strategies, iteratively refining its internal representations until a breakthrough is achieved. Unlike binary feedback systems, where only correctness is considered, CLU utilizes more granular insights from each incorrect iteration, which significantly accelerates convergence once a partial understanding is established. After the initial jump in accuracy, the system stabilizes near $100\%$, demonstrating that CLU has effectively internalized the transformation rule. Minor stochastic fluctuations in the performance can be attributed to the inherent variability within the underlying reasoning agent (\code{GPT-4o-mini}), allowing the system to explore slight variations around the learned rule, ultimately maintaining a stable equilibrium with reinforced knowledge.

These dynamics can be analogized to the cost landscape commonly observed in deep learning models, where the training trajectory initially might traverse a flat gradient, representing the absence of concrete or valuable knowledge, before descending rapidly into a steep minimum as the solution becomes apparent. In this context, the evolving knowledge spaces within CLU play a role similar to that of weights in a neural network, gradually adjusting and accumulating pertinent information that guides the system toward the optimal solution. During the early iterations, CLU behaves similarly to an NN initialized with random parameters, struggling to make meaningful progress due to the complex cost landscape, indicative of minimal or incorrect initial knowledge. As CLU accumulates valuable insights, it eventually finds a direction for improvement, akin to a network discovering a favorable gradient path toward the solution. The shape and profile of this learning curve depend heavily on the sophistication of the underlying reasoning agent: a more capable reasoner will have a more favorable cost landscape, leading to faster convergence, whereas a weaker agent faces a flatter landscape, potentially never reaching the solution. This analogy extends further when considering the effect of injecting prior knowledge, just as pre-trained weights in a DNN position it closer to an optimal solution, injecting hints or partial knowledge into CLU’s knowledge spaces can effectively place it near or at the solution's minima. For instance, adding abstract guidance like ``consider positional dependence'' primes the system, reducing the need for extensive exploration and accelerating the learning process. Moreover, if CLU starts with explicit instructions such as ``extract the second letter from each word,'' it is akin to initializing the network at or near the optimal solution, leading to immediate high accuracy, assuming the base reasoning engine has a minimal threshold of reasoning ability. 

CLU enables the systematic validation of our hypotheses about the learning process by facilitating direct interaction with its evolving knowledge base. As discussed in the \cref{sec:introduction}, CLU, though operating as a black box of extracted knowledge, uniquely provides the ability to interact with its internal state in both structured and unstructured ways, enabling a deeper understanding of its iterative learning dynamics. This interactive capability not only allows us to track CLU’s reasoning process in real-time but also enhances its explainability\cite{xu2019explainable} by providing transparent insights into how its knowledge evolves over time, offering an explicit view into what the system has learned and how it applies that knowledge. Specifically, every five iterations, along with running the tests for accuracy in \textit{reasoning phase}, we posed an explicit question to CLU: \code{What is the hidden transformation rule that you have learned?}. \Cref{fig:results}(c) illustrates the evolution of this internal understanding for the 1-shot setting. Initially, after the first iteration, CLU's reasoning is vague, such as ``The hidden transformation rule involves selecting specific letters from words in a phrase.'' This general understanding persists for several iterations, eventually evolving into more nuanced hypotheses, like ``The hidden transformation rule prioritizes vowels and significant consonants based on their sequential arrangement in words.'' Before reaching the inflection point, CLU refines its internal knowledge further to statements such as ``The hidden transformation rule considers the significance of letter position and type, such as distinguishing between nouns, verbs, consonants, and vowels.'' Finally, upon accumulating sufficient insights, CLU accurately identifies the transformation rule: ``The hidden transformation rule is to pick the second letter of each word,'' which corresponds to the observed jump in accuracy. This gradual evolution underscores CLU's capacity for iterative learning, where initial hypotheses are continuously refined based on feedback. Notably, GPT-4o-mini, even when used as part of the CLU framework, shows a tendency to favor reasoning involving vowels and consonants in early attempts—an observation consistent with its behavior in the IO and CoT setups. The strength of CLU lies in its reinforced feedback mechanism, which retains a memory of past failures and successful steps, thereby enabling it to a constant state of \textit{learning from blunders while solidifying breakthroughs}.

In \cref{fig:results}(b), we also illustrate the impact of varying the number of $n$-shot examples on the learning trajectory of CLU. Increasing the number of provided examples leads to an acceleration in the learning process, as evidenced by the earlier onset of the inflection point in accuracy. This behavior can be attributed to CLU's mechanism of accumulating knowledge from both past mistakes and successful attempts. With each additional reasoning attempt in every iteration, the Operational Agent gains access to a broader range of scenarios to explore, guided by the primary goal $G_m$. This enriched set of experiences enables CLU to refine its approach more swiftly and efficiently, extracting effective strategies from the provided examples.

To understand how \methodnameshort\ operates at a high level for this reasoning problem, we observe that it starts by attempting a variety of strategies akin to the process of intelligent exploration described in the introduction. Initially, it makes numerous mistakes, but the crucial factor in achieving meaningful progress is CLU’s ability to learn effectively from these mistakes, necessitating a robust memory of both errors and successes. There are different ways to structure this learning process: one approach, as adopted by CLU, is to disentangle the memory from the reasoning engine. This separation allows the knowledge base to be dynamically adjusted or pruned as needed, providing flexibility in the learning process. In contrast, the recent work by \citet{snell2024scaling} and OpenAI's \code{o1}\cite{OpenAI2024LearningToReason} take a different approach by holding memory directly in the context and generating reasoning progressively. In these studies, the system leverages past incorrect responses, effectively maintaining an ephemeral ``memory'' of its attempts within the context. By iteratively refining the trajectory of answer tokens—moving from errors to corrections—it learns to adjust its reasoning in response to past mistakes. This approach allows the model to correlate the correct answer with prior errors, implicitly identifying and correcting mistakes rather than ignoring earlier attempts entirely. Although o1's exact mechanism is proprietary and unknown, the general approach involves training the model to refine reasoning trajectories using in-context examples with methods like RL\cite{kumar2024training}. This method, however, has two significant disadvantages: the evolution of reasoning via an in-context setup proceeds autoregressively, which means there is no mechanism for dynamic modification or pruning—a critical step for learning, as seen from our method. Additionally, the context length imposes an inherent limitation on the amount of information that can be retained for reasoning. These limitations in the \citet{snell2024scaling} approach are counteracted, to some extent, through mechanisms like parallel exploration of various trajectories to add a degree of dynamism and fundamentally training the reasoner (foundational model) to better choose promising trajectories among multiple paths using curated examples. Nevertheless, these measures cannot entirely replicate the flexibility offered by CLU's disentangled architecture, where reasoning and memory evolve independently, enabling adaptive pruning and modification of accumulated knowledge without the theoretical limitation of the memory the system can hold. 

Although \code{GPT-4o-mini} consistently fails to decipher the transformation rule, OpenAI's \code{o1-preview}, with its use of episodic memory (though ephemeral), demonstrates the ability to try multiple patterns iteratively and ultimately derive the correct rule as shown in \cref{app:o1}. This memory-driven iterative reasoning approach is not orthogonal to CLU's methodology; in fact, in theory, the base reasoning engine in CLU could be swapped with models like \code{o1-preview}. Such integration would likely serve to augment the reasoning capabilities of CLU further, combining the strengths of dynamic memory management with powerful episodic reasoning.

\section{Conclusion}\label{sec:conclusion}

In this work, we introduced the \methodname\ (\methodnameshort), a framework designed for dynamic, feedback-driven learning that moves beyond traditional parameter optimization approaches. By iteratively refining both general and task-specific knowledge spaces, CLU has demonstrated a unique capacity for continual adaptation, successfully extracting abstract patterns in challenging scenarios that conventional LLMs struggled to solve. Unlike static deep learning models, CLU continuously incorporates feedback in real time, progressively enhancing its understanding and performance through iterative reasoning. However, the current serial nature of the learning process presents a key limitation, highlighting opportunities for optimization with cleverer learning routines and parallelization strategies. Additionally, future work could explore various directions, such as implementing more nuanced knowledge retrieval\cite{edge2024local,sarmah2024hybridrag} mechanisms, integrating modular tools, or even replacing the Operational Agent with more complex, goal-specific multi-agent systems, given CLU's inherent modularity. This modularity also allows CLU to be tailored to focused applications, where elements like the KMU could be augmented with episodic or hierarchical memory for more structured querying, further enhancing its adaptability.

Moreover, Composite Learning Systems comprising multiple CLUs represent an active area of extension, offering the potential to tackle even more sophisticated tasks through collaborative learning and shared knowledge refinement. The broader vision of CLU aligns with the growing consensus in the community on scaling inference compute\cite{kumar2024training,snell2024scaling,bansal2024smaller} to enhance reasoning capabilities while contributing further by integrating composite systems for greater adaptability and cooperation. This adaptability enables CLU to address a wide range of challenges, such as continual learning, knowledge distillation, and personalized AI—tasks that traditionally require specialized machine learning architectures—while also opening the door to many other unexplored applications. These areas represent just a few examples, and future work will continue to explore and expand CLU's capabilities, demonstrating its versatility for advancing complex real-world applications.

Our system combines basic reasoning agents with the capacity to actively learn from world knowledge. This integration, much like reasoning augmentation techniques such as Chain-of-Thought, enhances reasoning abilities but with an added dimension: the ability to learn from both past mistakes and successes. Whether this approach ultimately leads to scalable reasoning remains an open question that warrants further exploration.

\newpage
\bibliographystyle{plainnat}
\bibliography{main}
\newpage
\onecolumn
\appendix
\section{Iterative Reasoning in O1:}\label{app:o1}
\begin{figure*}[htb]
\begin{mdframed}[
    linecolor=bordercolor,
    linewidth=1pt,
    topline=true,
    rightline=true,
    leftline=true,
    bottomline=true,
    innertopmargin=5pt,
    innerbottommargin=5pt,
    innerrightmargin=5pt,
    innerleftmargin=5pt,
    backgroundcolor=white
]
    \begin{center}
        \textcolor{headingcolor}{\large\textbf{OpenAI \code{o1-preview} Reasoning Process}}
    \end{center}
    
    \vspace{5pt}
    
    \small
    \setlength{\parskip}{3pt}
    
    \textcolor{subheadingcolor}{\textbf{Analyzing Sentence Patterns}} \\
    \textcolor{textcolor}{I'll figure out if each word in the sentences links to a letter or code. Starting with evaluating word lengths and examining the given codes should illuminate the connection.}
    \vspace{2pt}
    
    \textcolor{subheadingcolor}{\textbf{Analyzing Letter Codes}} \\
    \textcolor{textcolor}{I'm noticing that each code is six letters long, and the length matches the number of words per sentence. I'm curious about the link between initial letters and the code.}
    \vspace{2pt}
    
    \textcolor{subheadingcolor}{\textbf{Mapping Letters}} \\
    \textcolor{textcolor}{I'm thinking through various ways to link the code with sentence letters. This involves summing alphabet positions, shifting, rearranging, or selecting specific letters based on word lengths.}
    \vspace{2pt}
    
    \textcolor{subheadingcolor}{\textbf{Testing Letter Patterns}} \\
    \textcolor{textcolor}{I'm analyzing the third letter of each word in a sentence to identify a pattern, but it doesn't match the expected code word.}
    \vspace{2pt}
    
    \textcolor{subheadingcolor}{\textbf{Trying Another Way}} \\
    \textcolor{textcolor}{I'm gathering info on using the first letters of each word to form a code, but they don't align. Now, I'm mapping letters to their position in the alphabet for new possibilities.}
    \vspace{2pt}
    
    \textcolor{subheadingcolor}{\textbf{Piecing Together Clues}} \\
    \textcolor{textcolor}{I'm gathering details from six words, calculating letter positions, and mapping a code to each word's letter sum. This involves modular arithmetic and examining the nth letter based on word length.}
    \vspace{2pt}
    
    \textcolor{subheadingcolor}{\textbf{Testing Letter Patterns}} \\
    \textcolor{textcolor}{I'm thinking through the initial letters of words from sentences. Using modulus operations to pinpoint specific letters, I uncovered 't', 'n', 'p', 's', 'n', 's'. This isn't yielding a match with 'rupcoi'.}
    
    \textcolor{subheadingcolor}{\textbf{Examining Letter Patterns}} \\
    \textcolor{textcolor}{I'm curious about the initial letters and their positions in the alphabet, calculating various sums like 54 and 82. This might hint at a cipher or letter mapping strategy.}
    \vspace{2pt}
    
    \textcolor{subheadingcolor}{\textbf{Mapping Transformations}} \\
    \textcolor{textcolor}{I'm aligning each word's initial letter with its code letter, calculating the difference in their positions in the alphabet to reveal hidden messages.}
    \vspace{2pt}
    
    \textcolor{subheadingcolor}{\textbf{Analyzing Letter Differences}} \\
    \textcolor{textcolor}{I'm working through mapping code letters to sentence words. Interestingly, there isn't an obvious pattern in the differences between code and sentence letter positions.}
    \vspace{2pt}
    
    \textcolor{subheadingcolor}{\textbf{Mapping Code Letters}} \\
    \textcolor{textcolor}{I matched each code letter to the second letter of the corresponding word, forming 'rupcoi'. This pattern was tested with the second sentence to ensure consistency.}
    \vspace{2pt}
\end{mdframed}

\caption{\textbf{OpenAI \code{o1-preview} Reasoning Process:} Example of OpenAI o1's reasoning process for analyzing and attempting to decode sentence patterns. The stepwise breakdown shows how o1 approaches the problem by evaluating, mapping, and iteratively refining its understanding. }
\label{fig:o1_reasoning_process}
\end{figure*}

The following appendix provides an illustrative example of the reasoning process employed by OpenAI's \code{o1-preview}\cite{openai_o1_preview} model on the same cryptographic encoding task described earlier. It is important to note that, as a commercial model, the exact internal mechanisms of \code{o1-preview} remain proprietary and are not publicly disclosed. However, it currently represents the state-of-the-art in iterative reasoning. The detailed breakdown presented here showcases how \code{o1-preview} iteratively refines its reasoning steps by exploring multiple potential patterns, leveraging episodic memory within a single inference context. This trajectory, exposed by OpenAI for transparency in evaluating the model's reasoning process, demonstrates how the model attempts various hypotheses and incrementally corrects its approach until it converges on the correct solution. Unlike the Composite Learning Unit (CLU), which disentangles memory and reasoning, \code{o1-preview} keeps an ephemeral memory within its inference context, thereby continuously refining its reasoning to arrive at a solution.

\end{document}